\pdfoutput=1
\documentclass[10pt,twocolumn,letterpaper]{article}

\usepackage{iccv}
\usepackage{times}
\usepackage{epsfig}
\usepackage{graphicx}
\usepackage{amsmath}
\usepackage{amssymb}

\usepackage[font=small,labelfont=bf]{caption}
\usepackage{shortbold}

\usepackage{booktabs}
\usepackage{gensymb}
\usepackage[accsupp]{axessibility} 

\usepackage[pagebackref=true,breaklinks=true,letterpaper=true,colorlinks,bookmarks=false]{hyperref}

\iccvfinalcopy

\makeindex
\begin{document}

\title{PixelSynth: Generating a 3D-Consistent Experience from a Single Image}

\author{Chris Rockwell\\
\and
David F. Fouhey\\
University of Michigan\\
\and
Justin Johnson\\
}

\maketitle

\begin{abstract}
  Recent advancements in differentiable rendering and 3D reasoning have driven exciting results in novel view synthesis from a single image.
  Despite realistic results, methods are limited to relatively small view change.
  In order to synthesize immersive scenes, models must also be able to extrapolate.
  We present an approach that fuses 3D reasoning with autoregressive modeling to outpaint large view changes in a 3D-consistent manner, enabling scene synthesis.
  We demonstrate considerable improvement in single-image large-angle view synthesis results compared to a variety of methods and possible variants across simulated and real datasets.
  In addition, we show increased 3D consistency compared to alternative accumulation methods.
\end{abstract}

\section{Introduction}

\begin{figure}[t!]
  \centering
  \includegraphics[width=\linewidth]{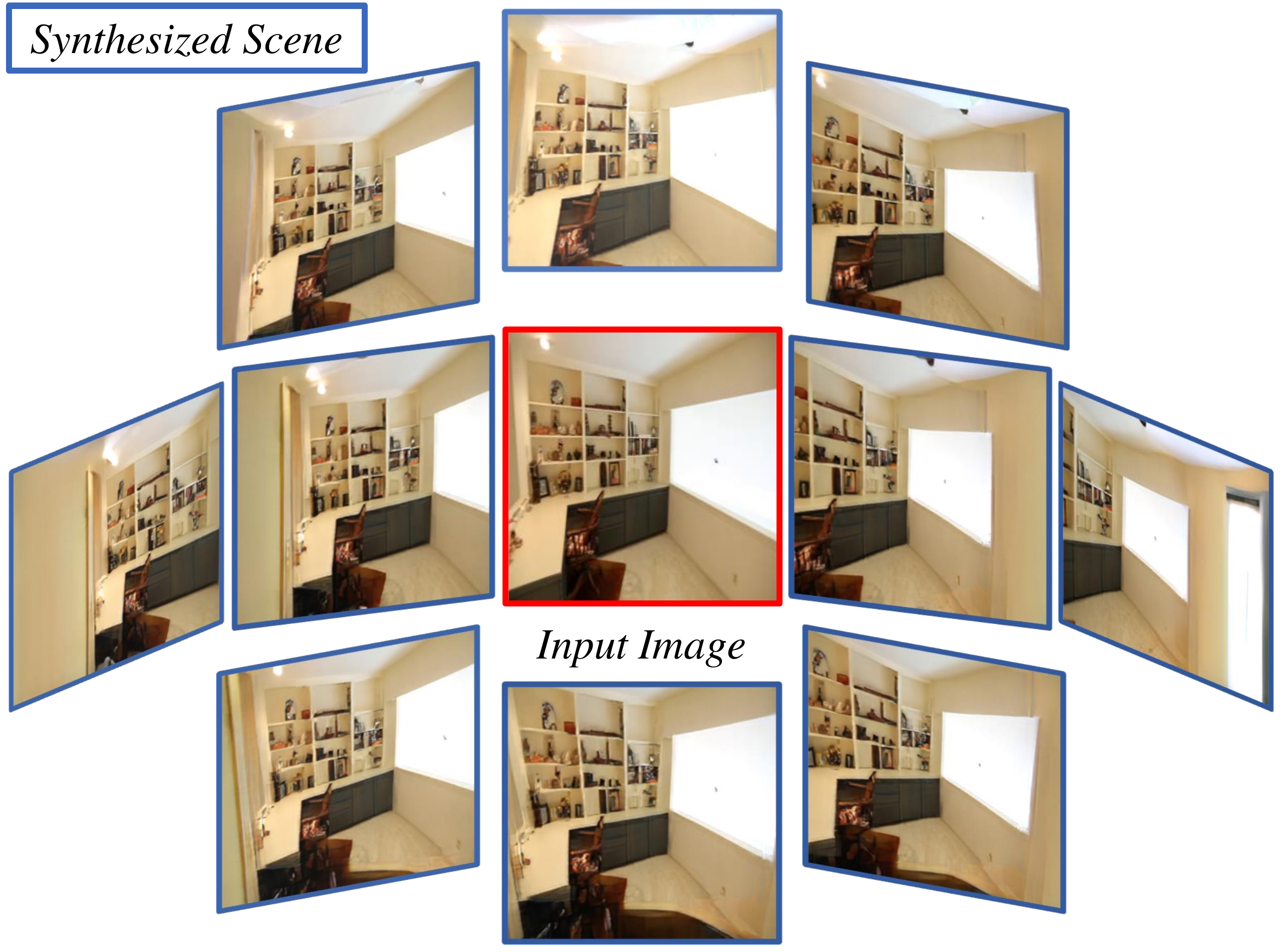}
  \caption{
    {\bf Single-Image Scene Synthesis.}
    Our framework fuses the complementary strengths of 3D reasoning and autoregressive modeling to create an immersive scene from a single image.
  }
  \label{fig:fig1}
\end{figure}

Imagine that you walk into the office shown in Figure~\ref{fig:fig1}.
What will you see if you turn right?
Is there a door onto a patio?
What if you step backward then look left?
While the image itself does not contain this information,
you can imagine a rich world behind the image due to your experience of other rooms.
This task of \emph{single-image scene synthesis} promises to bring arbitrary photos to life,
but requires solving several key challenges.
First, handling large view changes involves \emph{extrapolation} far beyond the input pixels.
Second, generating multiple outputs from the same input requires \emph{consistency}:
turning left by $10^\circ$ or $20^\circ$ should reveal progressively more of a single underlying world.
Finally, modeling view changes requires \emph{3D-awareness} to properly capture perspective changes.

Prior methods for view synthesis fall short of these goals.
There has been great progress at \emph{interpolating} between many input views of a single
scene~\cite{mildenhall2020nerf, meshry2019neural, shih20203d, sitzmann2019deepvoxels, tulsiani2018layer, wang2021ibrnet};
while these 3D-aware methods generate consistent outputs, they do not attempt to extrapolate
beyond their input views.
Prior approaches to single-image view synthesis~\cite{hu2020worldsheet, single_view_mpi, wiles2020synsin}
can extrapolate to small rotations and translations, but fail to model viewpoint changes at this scale.
For example, we show that na\"{i}vely retraining SynSin \cite{wiles2020synsin} for larger angles leads to collapse.

In parallel, autoregressive models have shown impressive results for image generation and
completion~\cite{chen2020generative, oord2016pixel, razavi2019generating, Salimans2017PixeCNN, van2016conditional}.
These methods are very successful at extrapolating far beyond the boundaries of an input image;
however they make no attempt to explicitly model a consistent 3D world behind their generated images.

In this paper we present an approach for single-image scene synthesis that addresses these challenges
by fusing the complementary strengths of 3D reasoning and autoregressive modeling.
We achieve \emph{extrapolation} using an autoregressive model to complete images when faced with large view changes.
Generating all output views independently would give inconsistent outputs.
Instead, we identify a \emph{support set} at the extremes of views to be generated (shown at the boundaries of Figure~\ref{fig:fig1}).
Generated images for the support set are then lifted to 3D and added to a consistent scene representation.
Intermediate views can then be re-rendered from the scene representation instead of generated from scratch,
ensuring \emph{consistency} among all outputs.

Producing a system that can both do extreme view synthesis and lift the results to 3D without requiring auxiliary data poses a challenge. 
Our approach, described in Section~\ref{sec:approach}, builds upon insights from both the view synthesis and autoregressive modeling communities.
Each image and new viewpoint yields a large and custom region to be filled in, which we approach by adapting VQVAE2 ~\cite{razavi2019generating} and expanding Locally Masked Convolutions \cite{jain2020locally} to learn image-specific orderings for outpainting.
Once filled, we obtain 3D using techniques from SynSin~\cite{wiles2020synsin}.
This system can outpaint large and diverse regions, accumulate outpainting in 3D, and can be trained without any supervision beyond images and 6DOF relative pose. 

We evaluate our approach as well as a variety of alternative methods and competing
approaches on standard datasets, Matterport 3D+Habitat \cite{Matterport3D,savva2019habitat} and RealEstate10K \cite{zhou2018stereo},
using substantially larger angle changes ($6 \times$ larger than \cite{wiles2020synsin}). 
Throughout the experiments in Section~\ref{sec:experiments}, we evaluate
with the standard metrics of human judgments, PSNR, Perceptual Similarity, and
FID. Our experimental results suggest that: (1) Our proposed approach produces
meaningfully better results compared to training existing methods on our larger viewpoints. In particular, users select our approach 73\% of the time vs. the best variant of multiple SynSin ablations. (2) Our approach of re-rendering
support sets outperforms alternate iterative approaches, being more consistent an average of 72\% of images.

\section{Related Work}

Both novel view synthesis and image completion have recently seen rapid progress. While novel view synthesis work has approached large view change,
it typically uses multiple input images. Given only a single input image, completion becomes highly relevant for outpainting.

\par \noindent {\bf Novel view synthesis.} If multiple views are available as input,
3D information can be inferred to synthesize new viewpoints.
Classical methods often use multi-view geometry \cite{chen1993view, debevec1996modeling, gortler1996lumigraph, levoy1996light, 1640800, zitnick2004high}.
Deep networks use a learned approach, and have shown impressive results with fewer
input views and less additional information.
They represent 3D in a variety of ways including depth images \cite{aliev2020, meshry2019neural, riegler2020free, yoon2020novel},
 multi-plane images \cite{single_view_mpi, zhou2018stereo},
point clouds \cite{wiles2020synsin}, voxels \cite{liu2020neural, sitzmann2019deepvoxels}, meshes \cite{hu2020worldsheet, kato2018neural, liu2019soft}, multi-layer meshes \cite{hedman2018instant, shih20203d}, and radiance fields \cite{mildenhall2020nerf, wang2021ibrnet, zhang2020nerf++}. We use a point cloud representation.

Given only a single input image,
CNNs have also achieved success \cite{chen2019monocular, liu2020infinite, tulsiani2018layer, yu2020pixelnerf}, owing largely to progress in generative modeling \cite{brock2018large, chen2017photographic, isola2017image, karras2020analyzing, ledig2017photo, park2019semantic, wang2018high, zhang2019self}. 
Nevertheless, single-image work has been limited to small angle change \cite{tatarchenko2016multi, single_view_mpi, zhou2016view}.
Methods such as SynSin \cite{wiles2020synsin} treat outpainting the same as inpainting, which struggles beyond a margin.
Our goal is to synthesize a scene from an image, which requires large outpainting.
We thus outpaint explicitly using a completion-based approach.

Concurrent work approaches similar, though distinct, problems.
Liu \textit{et al.} \cite{liu2020infinite} move forward on camera trajectories through nature scenes. 
In contrast, our focus is on indoor scenes and we handle outpainting. We show in this setting, a completion-based approach produces better results than a similar approach to \cite{liu2020infinite}.
Hu and Pathak \cite{hu2020worldsheet} use a mesh representation which undertakes twice the rotation of SynSin, but still treats outpainting as interpolation.
In contrast, our completion-based approach outpaints explicitly, which we show beats inpainting-based methods in large angles.
Rombach \textit{et al.} \cite{rombach2021geometry} approach large angle change on images, but do not learn a 3D representation. 
On scenes, we show our approach of accumulating 3D information across views is critical for consistency.

\vspace{1em}
\par \noindent {\bf Image Completion and Outpainting.} 
Recent work in inpainting takes an adversarial approach
\cite{iizuka2017globally, liu2018image, yang2017high, yu2018generative},
and has been used in novel view synthesis refinement \cite{liu2020infinite, sitzmann2019deepvoxels, single_view_mpi, wiles2020synsin}. 
However, inpainting is not suitable for synthesizing large angle change, which yields large missing regions.
Methods targeting outpainting \cite{teterwak2019boundless, wang2019wide, yang2019very} 
improve extrapolation, but are not flexible to arbitrary missing regions that may occur in view synthesis. 

Our work adopts techniques from the literature on deep autoregressive models. 
These works use masking with RNNs \cite{oord2016pixel}, CNNs \cite{menick2018generating, oord2016pixel, reed2017parallel, Salimans2017PixeCNN, van2016conditional}, and Transformers \cite{chen2020generative} to predict individual pixels sequentially.
While sequential generation is slower than feed-forward methods, it enables flexible ordering and state-of-the-art performance \cite{chen2020generative, chen2018pixelsnail, menick2018generating, oord2017neural, razavi2019generating}.
Yet, autoregressive methods by themselves do not enable 3D consistent outpainting. Thus, we build upon this literature in conjunction with 3D view synthesis to produce a set of {\it 3D consistent} views. Making this 3D fusion work requires building upon several recent developments: we adapt the masked convolution approach of Jain \textit{et al.} \cite{jain2020locally} to handle custom, per-image, regions to outpaint; also, like VQVAE2 \cite{razavi2019generating} and Dall-E \cite{unpublished2021dalle}, we find that selecting from a set of completions aids realism.

\section{Approach} \label{sec:approach}

Our goal is to input a single image and synthesize a consistent set of images
showing the surrounding scene.  This requires generating high-quality images
even under large transformations, and ensuring 3D consistency in the results.
We propose an approach to this task which uses deep autoregressive modeling to
facilitate high-quality extrapolations in conjunction with 3D modeling to ensure consistency. 

\begin{figure}[t!]
	\centering
			{\includegraphics[width=\linewidth]{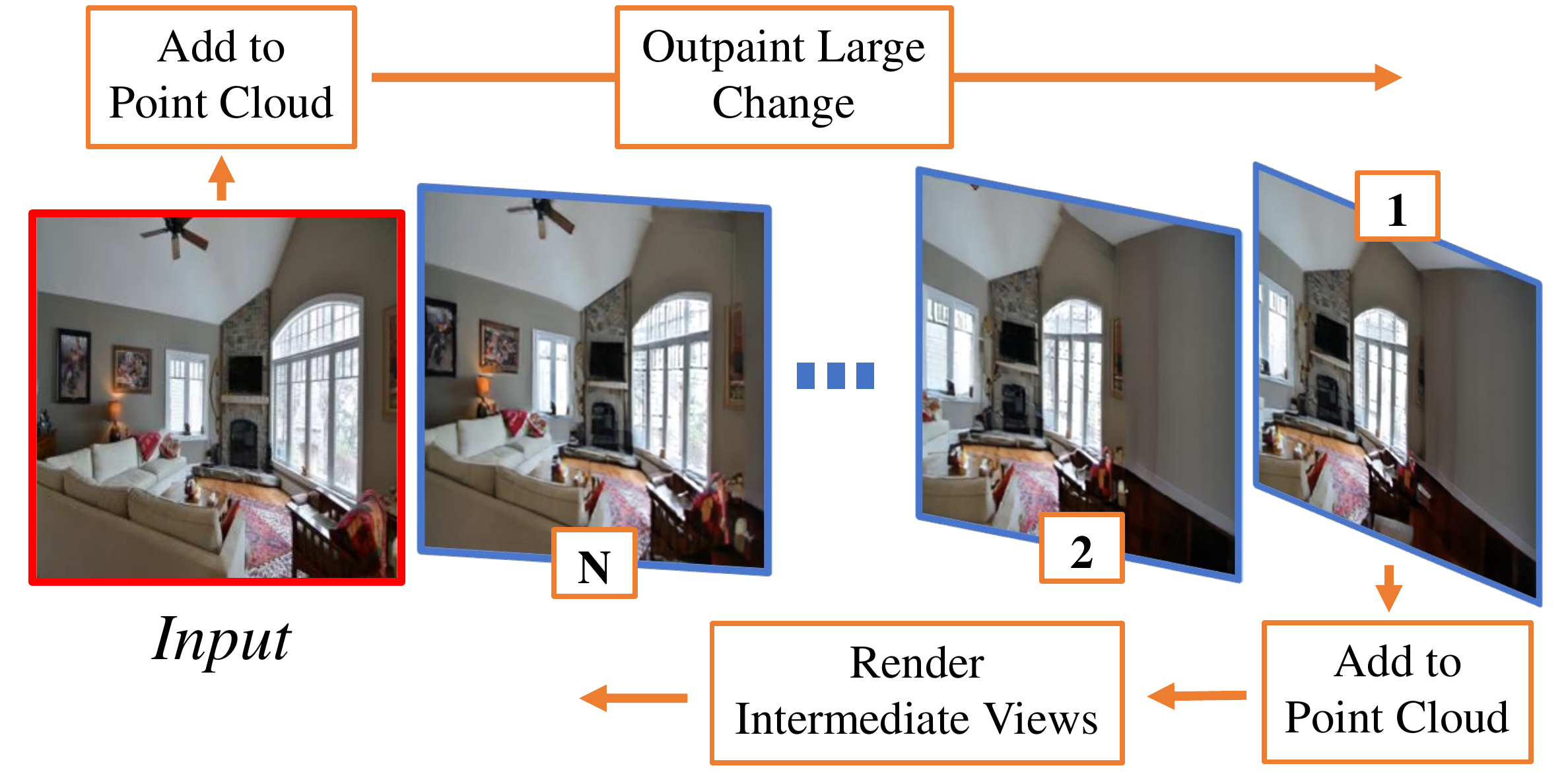}}
    \captionof{figure}{{\bf Consistent Scene Synthesis.} 
The model first generates extremal support views, from which intermediate views can be generated. This allows the model to outpaint once and re-render many times, which improves 3D consistency.}
	\label{fig:overview}
\end{figure}

The two critical insights of the method are the order in which 
image data is produced and the 3D nature of the approach. As illustrated
in Figure~\ref{fig:overview}, our system generates data on extremal
{\it support views} first and operates on point clouds. We outpaint these
support views with an autoregressive outpainting module that handles most of
the generation. As we reproject intermediate views, we touch up the results
with a refinement module. Throughout, we translate from images to point clouds
with a self-supervised depth module 
and back with a differentiable renderer.

\begin{figure*}[h!]
	\centering
			{\includegraphics[width=\textwidth]{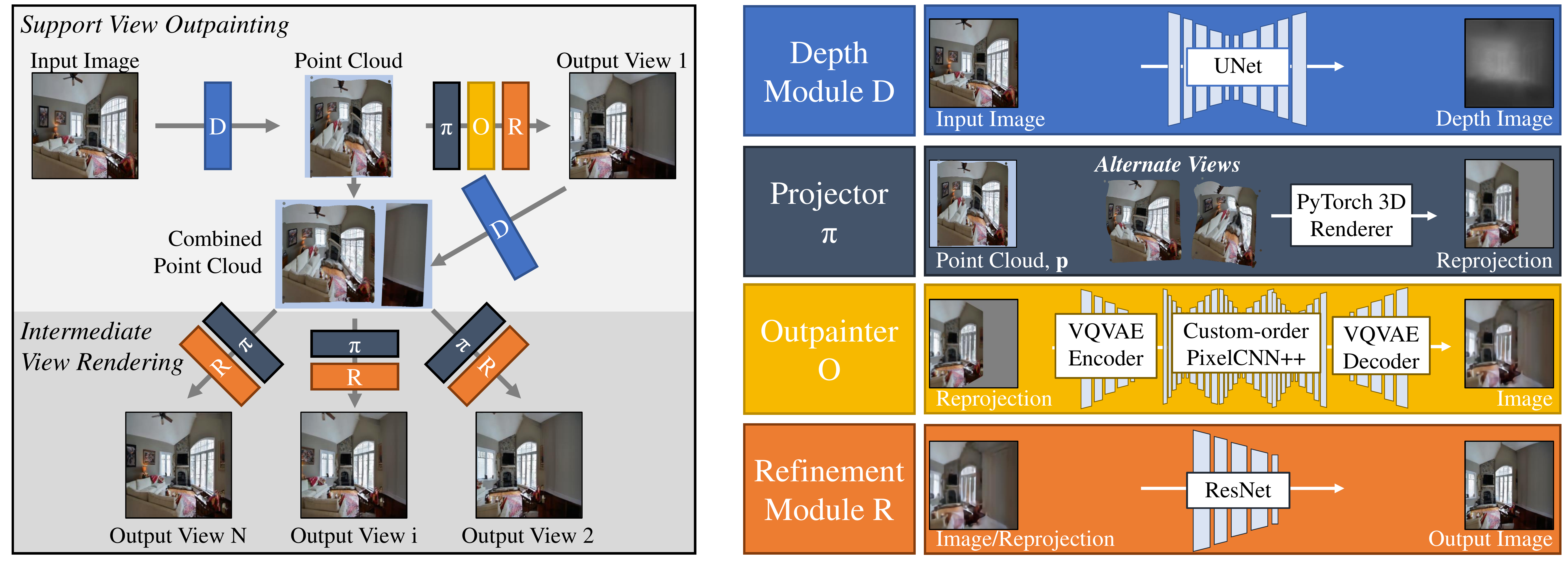}}
    \captionof{figure}{{\bf Approach Overview.} At inference, the model first outpaints an extremal support view,
    then renders intermediate views (left). Both steps rely on the Depth Module to lift images to point clouds, the Projector to render in a novel view, and the Refinement Module to smooth outputs (right). The Outpainter fills in missing information in the target view during outpainting.
}
	\label{fig:model}
\end{figure*}

\begin{figure}[t!]
	\centering
			{\includegraphics[width=\linewidth]{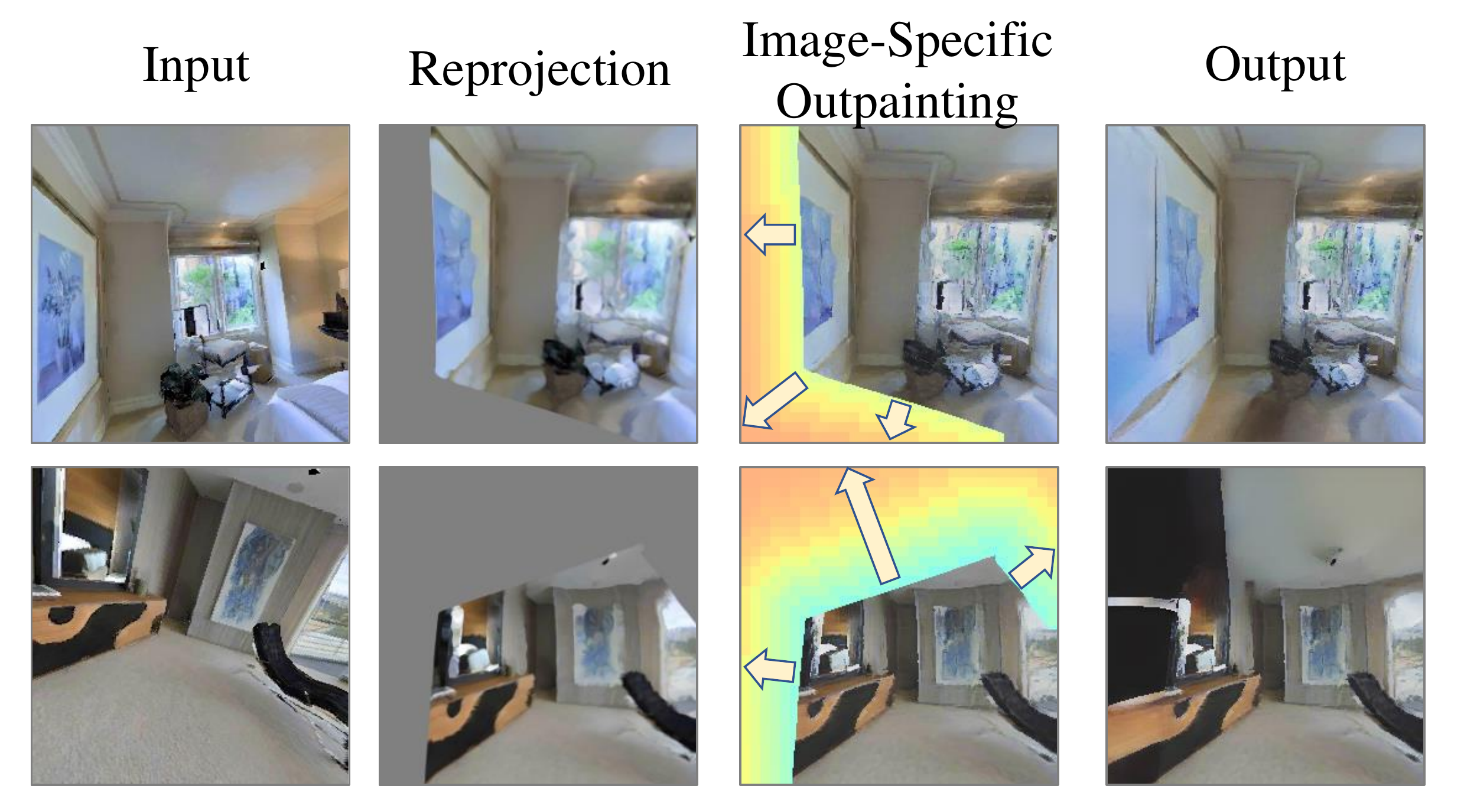}}
    \captionof{figure}{{\bf Autoregressive Outpainting.} We outpaint using image-specific orderings,
which begin with pixels adjacent to the visible region and move outward. Our model
outpaints a vector-quantized embedding space.}
	\label{fig:comp}
\end{figure}

\subsection{3D and Synthesis Modules}

We begin by introducing each of the modules used by our system, which are pictured on the right of Figure \ref{fig:model}.
We first describe the two modules that map to and from point clouds, followed by the models that generate
and refine pixels. With the exception of the projection module, all are learnable
functions that are represented by deep neural networks. 
Full descriptions of each are in the supplement.

\vspace{2mm}
\par \noindent {\bf Depth Module $D$:} Given an image $I$, we can convert it
into a colored point cloud $C = D(I)$ using a learned depth prediction system.
Specifically, the per-pixel depth is inferred using a U-Net \cite{ronneberger2015u} and the pixel
is mapped to 3D using known intrinsics. In our work, we learn $D$ via end-to-end training
on reprojection losses. 

\vspace{2mm}
\par \noindent {\bf Projector $\pi$:} Given a colored point cloud $C$, and
6DOF pose $\pB$, we can project it to an image $I = \pi(C,\pB)$ using Pytorch3D~\cite{ravi2020accelerating}'s
differentiable renderer. This renderer uses a soft z-buffer which blends nearby points.

\vspace{2mm}
\par \noindent {\bf Outpainter $O$:} When the viewpoint changes dramatically, large missing regions
come into the field of view and must be outpainted. The specific regions depend on both the viewpoint shift
and the image content. We perform per-image outpainting on the latent space of a 
VQ-VAE \cite{oord2017neural, razavi2019generating}. Our particular model autoencodes $256 \times 256 \times 3$ inputs through 
a discrete $32 \times 32 \times 1$-dimensional embedding space $\{Z\}$, $Z_{i,j,1} \in \mathbb{Z}^{512}_{1}$.
Using discrete values encourages the model to choose more divergent completions.

We outpaint in this $32 \times 32$ latent space using an autoregressive model ~\cite{oord2017neural,unpublished2021dalle,esser2020taming}. In our particular 
case we predict pixel embeddings with a PixelCNN++ \cite{Salimans2017PixeCNN} architecture,
using a $512$-way classification head to predict a distribution over embeddings. 
We use Locally Masked Convolution \cite{jain2020locally} blocks to implement image-specific custom-pixel orderings. 
We show examples of the orders used in Figure \ref{fig:comp}, which outpaint pixels close to the visible region, followed by those farther away.

\vspace{2mm}
\par \noindent {\bf Refinement Module $R$:} Outpainting returns images that are sensible, but often lack details or have inconsistencies due to imperfect depth. We
therefore use an adversarially-trained refinement module to correct local errors. This module 
blends the reprojection of the original and outpainted pixels and predicts a residual to its input. Our generator architecture is similar to \cite{wiles2020synsin} and uses 
8 ResNet \cite{he2016deep} blocks containing Batch Normalization
injected with noise like \cite{brock2018large}. We adopt the discriminator from \cite{wang2018high}.

\subsection{Inference}

At inference time, we compose the modules to generate the full set of images in 
two phases: support view outpainting,
followed by intermediate view rendering. The process overview is shown on the left of Figure \ref{fig:model},
and can be reused to synthesize multiple viewing directions.

\vspace{2mm}
\par \noindent {\bf Support View Outpainting and Refinement:} Given a single input image and
support view $\pB_1$, our goal is to create an updated point cloud that includes
a set of pixels that might be seen in view $\pB_1$. We achieve
this by outpainting in the support view: first, we estimate what can be
seen in the support view by projecting the point cloud inferred
from the input, or $I' = \pi(D(I),\pB_1)$. 
This projection usually has a large and image-specific gap (Figure \ref{fig:comp}). 
Our second step composes the outpainting and refinement module, or $I_1 = R(O(I'))$.
Finally, the resulting large-view synthesis itself is lifted to a point cloud by applying $D$, or $C_1 =D(I_1)$. 

\vspace{2mm}
\par \noindent {\bf Intermediate View Rendering and Refinement:} Once the input and outpainted support views have 
converted to point clouds, we can quickly render any intermediate view $\pB_i$ by applying the projection
and refinement modules. Specifically, if $C = D(I)$ is the input point cloud and $C_1$ is the support view point
cloud, we simply apply the refinement to their combined projection, or $R(\pi([C,C_1],\pB_i))$.

\subsection{Training and Implementation Details}

End-to-end training of the model is difficult because the Outpainter requires
ground truth input and output, which breaks gradient flow. We therefore perform training of
the Depth and Refinement Modules, the Outpainter VQ-VAE, and Outpainter autoregressive model separately.
In all cases, batch size is chosen to maximize GPU space, and training stops when validation loss plateaus.

We first train the latent VQ-VAE space of the Outpainter $O$, which is then frozen and used by the other modules during training.
Training takes $\sim$ 30k iterations with batch size of 120 and uses the losses from \cite{razavi2019generating}: 
an L\textsubscript{2} reprojection loss and an embedding commitment loss.

Next, we train the Depth and Refinement Modules jointly.
Ground truth is used in place of missing pixels to be outpainted to avoid
having to sample from the Outpainter during training. 
The composition of Depth and Refinement is trained
with an L\textsubscript{1} pixel loss, a content loss \cite{zhang2018unreasonable}, and
a multi-scale discriminator loss \cite{wang2018high}. The discriminator in the
Refinement Module is trained at multiple scales with a feature-matching loss.
In the process, the Depth Module
is implicitly learned. We train for 125k iterations (200k on Matterport) with batch size 12.

Finally, the autoregressive model in $O$ is trained. 
It is trained upon the learned VQ-VAE latent space using custom outpainting orderings.
Orders move outward from reprojections (Figure \ref{fig:comp}); reprojections are a function of depths predicted by the Depth Module. 
Training takes $\sim$ 75k iterations using a batch size of 60 and cross-entropy loss.

\par \noindent {\bf Curriculum Learning.}
The Depth and Refinement Modules are trained via a curriculum.
They first learn to synthesize small view changes, then generalize to larger angles.
For the first 25k iterations, we train at the same rotation as \cite{wiles2020synsin}. For Matterport, this is 20$\degree$ in each Euclidean direction, for RealEstate10K it is 5$\degree$ total.
Next, we increase maximum rotation by this amount, and repeat this increase every 25k iterations until reaching our target rotation.

\par \noindent {\bf Outpainting Inference Details:}
Outpainting produces diverse completions, which is a dual-edged sword:
some are good, but many will be inconsistent with the input.
We thus produce multiple samples and select the best.
Selection uses the complementary signals of classifier entropy and discriminator losses -- samples that are consistent with 
inputs usually have less entropy over model classes (we use a Places~\cite{zhou2017places} classifier),
and detailed images usually have higher discriminator loss.
Full details are in supplement.

\begin{figure*}[t!]
	\centering
			{\includegraphics[width=\textwidth]{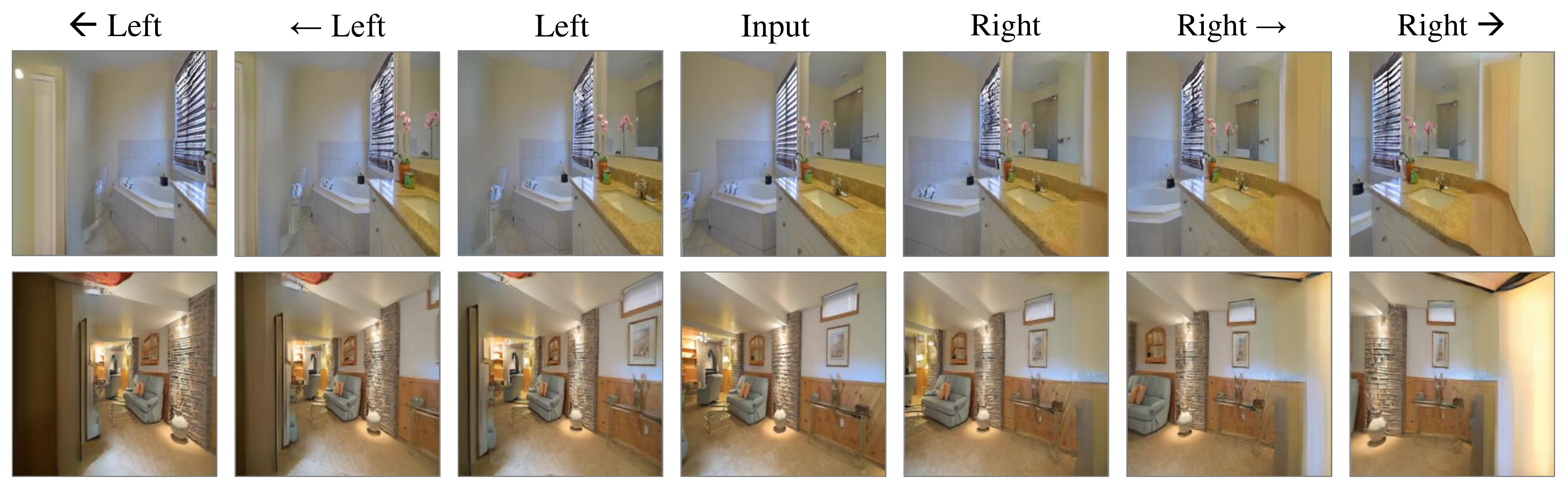}}
    \captionof{figure}{{\bf Consistent, High Quality Scenes.} Given a single image, the proposed method generates images across large viewpoint changes. It both continues content (e.g. wall, bottom right) and invents consistent content (e.g. door, top left). Results shown on RealEstate10K.}
	\label{fig:consistency}
\end{figure*}

\par \noindent {\bf Computational Reduction.} We perform aggressive yet efficient pruning to the aggregate model, which becomes heavy otherwise.
The Outpainter is most critical to speed.
We reduce the depth of the autoregressive model by 60\% and reduce width by
50\%, and use $32\times32$ completions compared to VQVAE2's of
both $32\times32$ and $64\times64$ completions.
We find that detail from $64\times64$ completions can instead be generated
by pairing a $32\times32$ completions with the refinement module.

In total, our changes improve our inference speed by $10\times$ for 50 completions ($500\times$ for one completion), compared to using a full VQVAE2 and PixelCNN++ setup. 
One completion takes $\approx$ 1 minute using 50 samples, or 1 second with 1 sample.
Training takes $\approx$ 5 days on 4 2080 Ti GPUs.

\section{Experiments} \label{sec:experiments}

The goal of our experiments is to identify how well our proposed method can
synthesize new \textit{scenes} from a single image. We do this on standard
datasets and compare with the state of the art (Sec.~\ref{sec:exp_setup}).
Our task requires not only creating plausible new content, but also ensuring the created content is
3D consistent. We evaluate these two goals separately. We test the generated views
for quality by independently evaluating each generated view (Sec.~\ref{sec:exp_quality});
we measure consistency by evaluating consistency across a {\it set} of overlapping 
views (Sec.~\ref{sec:exp_consistency}).

\begin{figure*}[t!]
	\centering
			{\includegraphics[width=\textwidth]{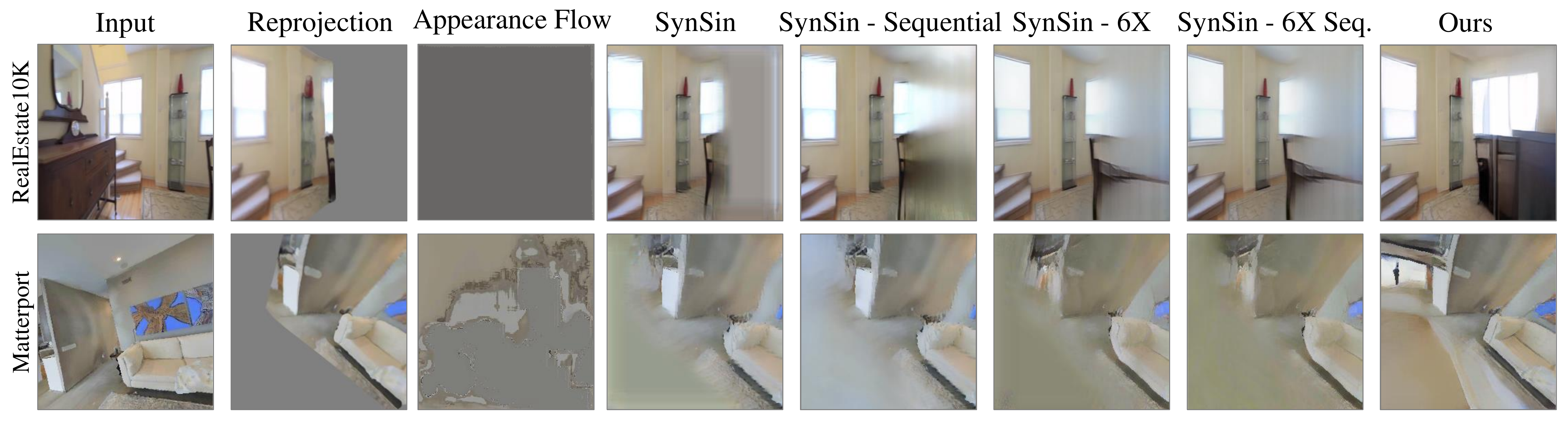}}
    \captionof{figure}{{\bf View Synthesis Ablations.} 
Prior work is not capable of synthesizing large angle change, even with additional training and sequential generation. 
This typically leads to collapse. Explicit outpainting instead creates realistic and consistent content.
}
	\label{fig:ablations}
\end{figure*}

\subsection{Experimental setup}
\label{sec:exp_setup}

We evaluate throughout on standard datasets, using standard metrics. We compare our approach with baselines from the state
of the art, as well as ablations that test alternate scene generation or view synthesis strategies.

\vspace{2mm}
\par \noindent {\bf Datasets.} 
Following \cite{wiles2020synsin}, we evaluate on Matterport3D \cite{Matterport3D} and RealEstate10K \cite{zhou2018stereo}. 
These enable the generation of pairs of views for training and evaluation.
For consistency with past work, we follow a similar selection setup as \cite{wiles2020synsin}, except we increase
rotations; making corresponding changes in sampling to do so.
Full details appear in the supplement.

\vspace{1mm}
\par \noindent {\it Matterport:} Image selection is done by an embodied agent
doing randomized navigation in Habitat~\cite{savva2019habitat}.  We increase
the limits of \cite{wiles2020synsin} angle selection from $20^\circ$ in each
direction to $120^\circ$.

\vspace{1mm}
\par \noindent {\it RealEstate10K:} RealEstate10K is a collection of videos and image collection consists
of selecting frames from a clip. SynSin selects pairs with angle changes of $\ge 5^\circ$ with maximum frame
difference of 30. Increasing the angle change is not straightforward since $\ge 30^\circ$ changes are infrequent
and can correspond to far away frames from different rooms. We therefore select pairs of between $20^\circ$ and $60^\circ$ 
apart and $\le 1$m away. The average angle is $\approx 30^\circ$, roughly $8\times$ larger than SynSin. 
SynSin average angle is less than $5^\circ$ because it sometimes re-samples; see \cite{wiles2020synsin} for details.

\vspace{2mm}
\par \noindent \textbf{Evaluation Metrics.} We evaluate content quality and
consistency using human judgments as well as a set of automated metrics.

\vspace{1mm}
\par \noindent \textit{Human A/B Judgments:} We evaluate image quality by asking annotators to compare generated images
and consistency by asking annotators to compare image pairs. In both cases, we ask humans to make pairwise comparisons
and report average preference rate compared to the proposed method: a method is
worse than the proposed method if it is below $50\%$. Automatic evaluation of
synthesis is known to be difficult, and we find that human judgments
correlate with our own judgments more than automated
systems.

\vspace{1mm}
\par \noindent \textit{Fr\'echet Inception Distance (FID) \cite{heusel2017gans}:} We evaluate how well generation images match on a distribution level
using FID, which measures similarity by comparing distributions of
activations from an Inception network. 
It has been shown to correlate well with human judgments \cite{heusel2017gans}, and we find it is the best automated measure of image quality. 

\vspace{1mm}
\par \noindent \textit{PSNR and Perceptual Similarity \cite{zhang2018unreasonable}:} PSNR and
Perc Sim are standard metrics for comparing images. They are excellent measures of
{\it consistency}, which is a unimodal task. Prior work 
\cite{teterwak2019boundless, wang2019wide, yang2017high} suggests that they are
poor measures for conditional image generation since there are many modes
of the output. We report them only for
consistency with past work.

\vspace{2mm}
\par \noindent \textbf{Baselines.} We compare with existing work in the space
of synthesizing unseen parts of rooms, as well as ablations that test
components of our system (which are introduced when used). Our primary point of comparison is SynSin
\cite{wiles2020synsin} since it is state of the art, although we evaluate other
standard baselines \cite{tatarchenko2016multi,single_view_mpi, zhou2016view}.
In addition to standard SynSin, we evaluate many approaches to extending SynSin
to handle the large rotations in our dataset.

\vspace{1mm}
\par \noindent \textit{SynSin \cite{wiles2020synsin}:} One primary baseline is SynSin as described in~\cite{wiles2020synsin} with no adaptation for extreme view change. We also evaluate the following
extensions: \textit{(SynSin - Sequential)} an autoregressive SynSin that breaks the transformation into 6 smaller transforms, accumulating 3D information;
\textit{(SynSin - 6X)} a SynSin model trained on larger view change;
\textit{(SynSin - 6X, Sequential)} a SynSin model trained on larger view changes and evaluated sequentially.

\vspace{1mm}
\par \noindent \textit{Other Baselines:} We compare with a number of other view synthesis approaches, which tests whether any difficulties are
specific to SynSin. In particular, we use: \textit{Appearance Flow \cite{zhou2016view}};
\textit{Tatarchenko \textit{et al.} (Multi-View 3D from Single Image) \cite{tatarchenko2016multi}}; and 
\textit{Single-View MPI \cite{single_view_mpi}}, which is only available on RealEstate10K.

\vspace{2mm}

\subsection{Evaluating Quality}
\label{sec:exp_quality}

\begin{table}[t!]
\caption{{\bf Image Quality} as measured
by A/B testing (preference frequency for a method compared to ours) as
well as FID.
In A/B tests, workers select the synthesized image better matching image reprojections
choosing from the alternate method and ours.
All baselines are preferred less often than our approach, and our approach
better matches the true distribution as measured
by FID.
Single-View MPI \cite{single_view_mpi} is not available on Matterport. 
}
\centering 
{
  \begin{tabular}{@{~}l@{~~~~~}c@{~}c@{~~~~~}c@{~}c@{~}}
    \toprule
    Method & \multicolumn{2}{c}{Matterport} & \multicolumn{2}{c}{RealEstate}\\

     & A/B $\uparrow$ & FID $\downarrow$ & A/B $\uparrow$ & FID $\downarrow$ \\
\midrule
Tatarchenko \textit{et al.} \cite{tatarchenko2016multi} & 0.0\% & 427.0 & 0.0\% & 256.6 \\
Appearance Flow \cite{zhou2016view} & 19.8\% & 95.8 & 1.9\% & 248.3 \\
Single-View MPI \cite{single_view_mpi} & - & - & 2.7\% & 74.8 \\
SynSin \cite{wiles2020synsin} & 14.8\% & 72.0 & 5.8\% & 34.7  \\
SynSin - Sequential & 19.5\% & 77.8 & 11.5\% & 34.9  \\
SynSin - 6X & 27.3\% & 70.4 & 22.0\%  & 27.9 \\
SynSin - 6X, Sequential & 21.2\% & 79.3 & 14.4\% & 33.1 \\
\midrule
Ours & - & 56.4 & - & 25.5   \\
\bottomrule
\end{tabular}
\vspace{-0.1in}
}
\label{tab:overview}
\end{table}

We begin by measuring the generated image quality.  Being able to synthesize realistic
images well beyond inputs is critical to generate an immersive scene.  

\vspace{2mm}
\par \noindent {\bf Qualitative Results.}
Figure \ref{fig:consistency} shows that the proposed method can produce high-quality, 3D-consistent images across large angle changes.
The images suggest the method is capable of continuing visible scene information realistically,
including an entirely new door that is consistent with the original image content on top left
and the continuation of the textured wall on bottom right.

Comparisons with prior work in Figure \ref{fig:ablations} show that the
baselines struggle with large angle changes. Straightforward
solutions like sequential generation or training on large angle changes do not succeed. While SynSin - 6X generates
some results, it mainly repeats visible pixels. Our approach can extend visible information where
appropriate, but also creates new objects like desks, windows, and tables.

\vspace{2mm}
\par \noindent {\bf Quantitative Results.} Quantitative
results in Table~\ref{tab:overview} are largely consistent with qualitative
results from Figures~\ref{fig:ablations}. On Matterport, our explicit outpainting does substantially better 
across metrics compared to baselines including SynSin.
Alternative baselines to SynSin perform worse, showing this is not a failing specific to SynSin. 
Training on larger rotation and applying sequential generation to SynSin help, but do not close the gap to our method.

On RealEstate10K, the gap is even larger for human judgment. 
Interestingly, SynSin - 6X does well on FID on RealEstate10K despite often producing repeated and mean colors. 
This is in part because RealEstate10K contains a high frequency of images looking through doorways.
In these cases, the target view often includes the wall next to the doorway, which typically consists of bland and repeated colors. 
Thus, repeated colors become reasonable at a distribution level, even if the difference is clear to humans.

To follow past work, we report PSNR and Perceptual Similarity metrics in Table \ref{tab:bad} for best performing methods (see supplement for all). 
These automated metrics, especially PSNR, are poor measures for extrapolation tasks \cite{teterwak2019boundless, wang2019wide, yang2017high}, so A/B testing is the primary measure of success.
The results of Appearance Flow in Figure \ref{fig:ablations} is evidence of this phenomenon.
This method often produces entirely gray images in RealEstate10K, and loses to our method 98.1\% of the time in A/B testing. 
Yet, its PSNR is competitive with other methods.

\subsection{Evaluating Consistency}
\label{sec:exp_consistency}

\begin{table}[t!]
\caption{Traditional metrics such as PSNR are poor measures for extrapolation tasks \cite{teterwak2019boundless, wang2019wide, yang2017high}, but are reported for reference.
}
\centering 
\resizebox{\ifdim\width>\linewidth
        \linewidth
      \else
        \width
      \fi}{!}
{
  \begin{tabular}{@{~}l@{~}c@{~}c@{~}c@{~}c@{~}}
    \toprule
    Method & \multicolumn{2}{c}{Matterport} & \multicolumn{2}{c}{RealEstate10K}\\
     & PSNR $\uparrow$ & Perc Sim $\downarrow$ & PSNR $\uparrow$ & Perc Sim $\downarrow$ \\
\midrule
Tatarchenko \textit{et al.} \cite{tatarchenko2016multi} & 13.72 & 3.82 & 10.63 & 3.98 \\
Appearance Flow \cite{zhou2016view} & 13.16  & 3.68 & 11.95 & 3.95 \\
Single-View MPI \cite{single_view_mpi} & - & - & 12.73 & 3.45 \\
SynSin - 6X, Sequential & 15.61 & 3.17 & 14.21 & 2.73 \\
\midrule
Ours & 14.60 & 3.17 & 13.10 & 2.88   \\
    \bottomrule
\end{tabular}
\vspace{-0.1in}
}
\label{tab:bad}
\end{table}

\begin{figure}[t!]
	\centering
			{\includegraphics[width=\linewidth]{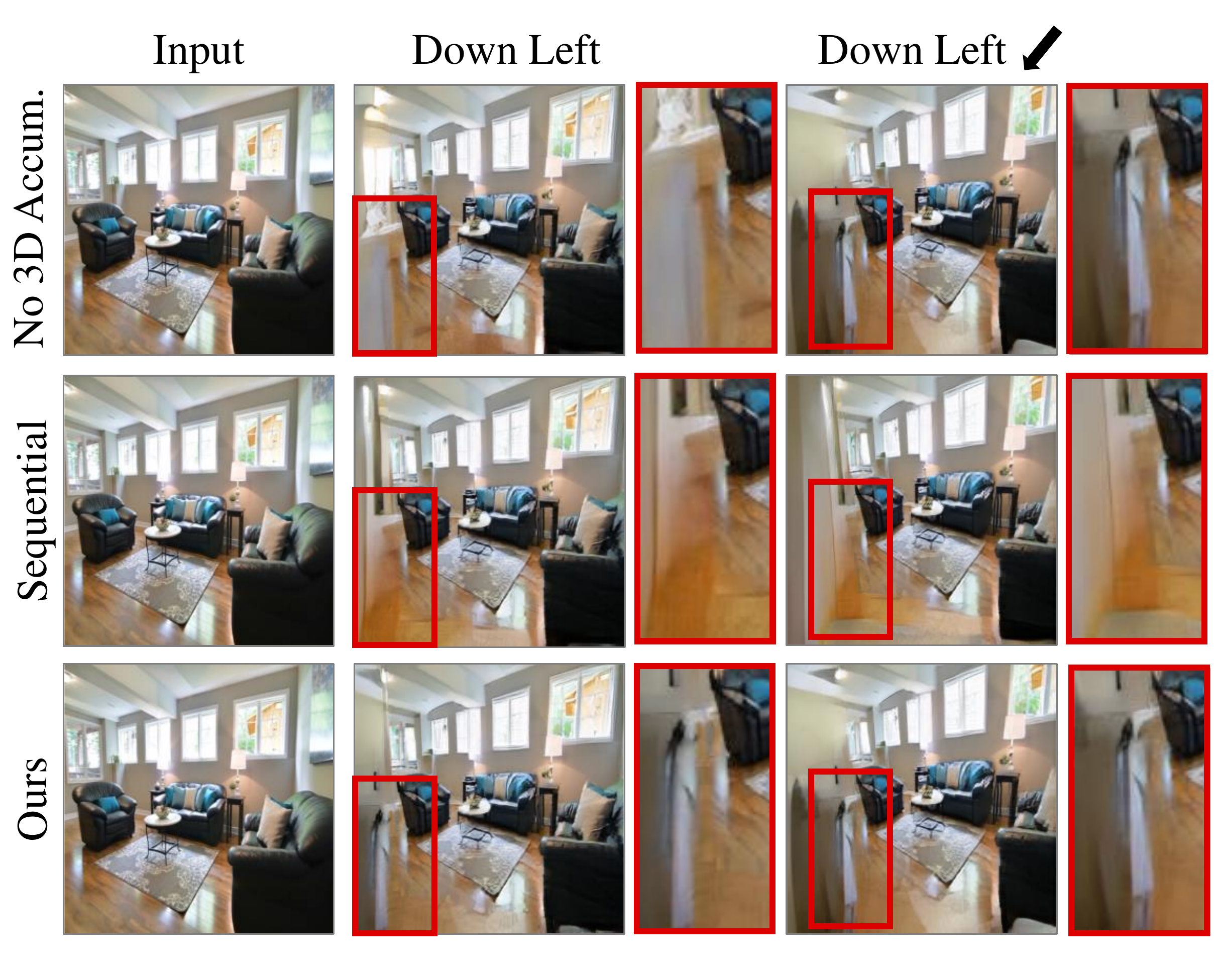}}
    \captionof{figure}{{\bf Consistency Ablations.} The proposed method generates a consistent scene across views. Without 3D accumulation, outpainted regions are completely inconsistent. Sequential outpainting yields artifacts, due to using autoregressive completions in multiple views.}
	\label{fig:consistency2}
\end{figure}

Having evaluated the quality of individual images, we next evaluate 
consistency. We note that consistency only matters if results
are of high quality -- producing a constant value is consistent.
We therefore focus only on our approach and
alternate accumulation strategies for our method. We 
evaluate consistency between a pair of generated results, one extreme
view and an intermediate view. The setup follows
view synthesis, with two exceptions:
we pick a large view change ($\sim$ $35^\circ$ horizontal, $\sim$ $17.5^\circ$ vertical) to ensure enough 
change to check consistency, and we use only horizontal and vertical rotation since
camera roll makes judging consistency difficult.
Full details are in supplement.

\vspace{2mm}
\par \noindent {\bf Alternate Strategies For Scene Synthesis.} Throughout, we use our base model but compare with alternate strategies for scene synthesis. Specifically, we try:

\vspace{1mm}
\par \noindent \textit{Ours - No 3D Accumulation:} We apply the method without accumulating the point cloud across generated images.
This means outpainting takes place for each synthesized view, and outpainting is independent across views.

\vspace{1mm}
\par \noindent \textit{Ours - Sequential Generation:} We apply the proposed 3D accumulation using the reverse order: this outpaints the missing region for the nearest image, then repeats outward.
This results in outpainting in each new view, compared to our method which outpaints only one extremal view.

\vspace{2mm}
\noindent {\bf Qualitative Results.} We show two outputs for an image in Figure~\ref{fig:consistency2}.
Without accumulation, one gets two wildly different results for each of the views (top row). Adding
the accumulation helps resolve this, but doing it sequentially in two stages (middle row) produces visible
artifacts. By generating a single large change first, our approach (bottom row) produces more consistent results.

\begin{table}[t!]
\caption{{\bf Scene Consistency}. A/B comparison of consistency. Workers select the most consistent pair of overlapping synthesized images (e.g. right two full images in Figure \ref{fig:consistency2}). All scores below 50 indicate the proposed method beats all ablations, on average. Consistency is lowest without 3D accumulation. Sequential order generation is less consistent than ours due to repeated outpainting.
}
\setlength{\tabcolsep}{5pt} 
\centering 
\resizebox{\ifdim\width>\columnwidth
        \columnwidth
      \else
        \width
      \fi}{!}
{
  \begin{tabular}{l c c c c}
    \toprule
     & \multicolumn{2}{c}{A/B vs. Ours $\uparrow$} \\
    Method & \multicolumn{1}{c}{Matterport} & \multicolumn{1}{c}{RealEstate10K} \\
\midrule
No 3D Accumulation & 22.6\% & 7.5\% \\ 
Sequential Generation & 44.0\% & 36.2\% \\ 
\midrule	
Ours & - & - \\
    \bottomrule
\end{tabular}
\vspace{-0.1in}
}
\label{tab:consistency}
\end{table}

\vspace{2mm}
\noindent {\bf Quantitative Results.} A/B testing shown in Table~\ref{tab:consistency} supports the qualitative findings:
On RealEstate10K, independent generation is chosen only 7.5\% of the time. 
Sequential generation accumulates a 3D representation, and performs better than the na\"{i}ve method, but is less consistent than the proposed method.

We quantitatively validate these results using PSNR and Perceptual Similarity in a controlled setting.
We use the same setup, but apply pure rotations to images, which means the resulting 
images are related by a homography. 
We apply this on RealEstate10K and use homographies to warp
extreme to intermediate views and to warp intermediate to extreme views.
We then calculate consistency using PSNR on overlapping regions and Perc Sim on warped images with non-overlapping regions masked.
Without 3D accumulation does poorly, with Perc Sim/PSNR 0.606/13.6; sequential generation with 3D accumulation improves results tremendously to
0.456/17.9. The full method improves further to 0.419/18.6.

\subsection{Ablations}

Finally, we report some ablations of the method. These test the contribution of our latent space, the use of 
multiple samples, and the mechanism used to select the samples.

\vspace{2mm}
\par \noindent {\bf Ablations.} We compare the proposed Outpainting Module and Sampling to alternatives.

\vspace{1mm}
\par \noindent \textit{Ours - RGB Autoregressive:} We compare with using a RGB space to test the value of our latent space.
Similar to prior work \cite{Salimans2017PixeCNN}, we only consider a single completion for RGB. As opposed to VQ-VAE-based models, multiple completions is less helpful empirically.

\vspace{1mm}
\par \noindent \textit{Ours - 1 Completion:} We evaluate effectiveness of our method with just one completion, which is more efficient but
often less effective.

\vspace{1mm}
\par \noindent \textit{Ours - Classifier Selection:} We apply our proposed method without using a discriminator for selection. 

\vspace{1mm}
\par \noindent \textit{Ours - Discriminator Selection:} We apply our proposed method without classifier included for selection.

\begin{figure}[t!]
	\centering
			{\includegraphics[width=\linewidth]{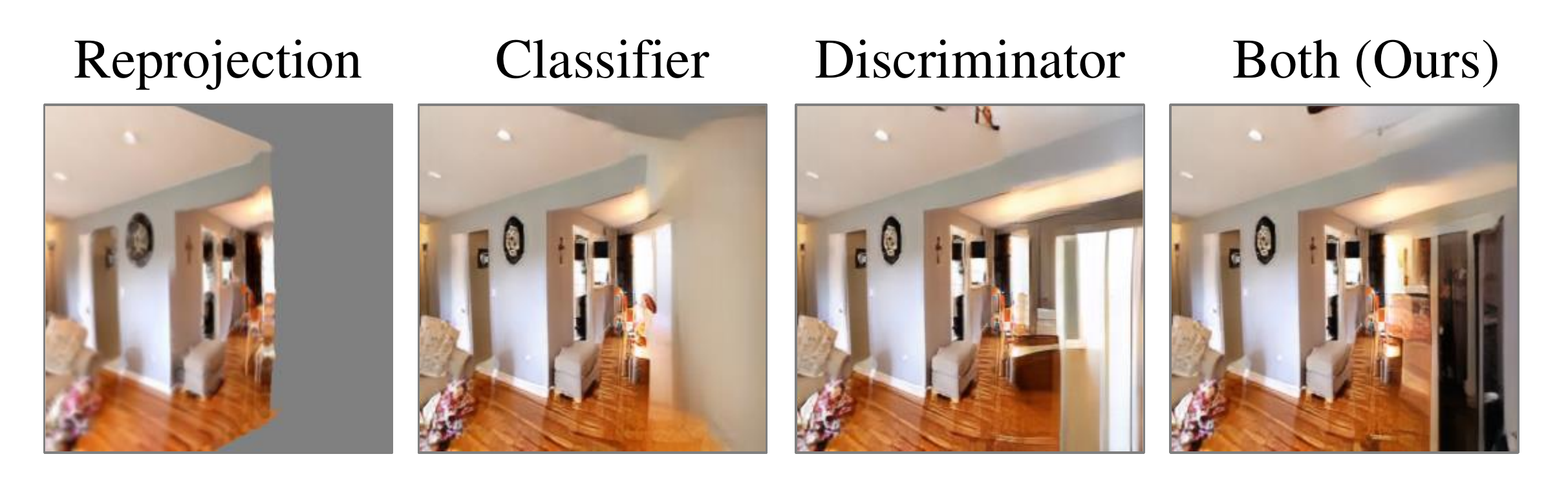}}
    \captionof{figure}{{\bf Improving Sample Selection.} Selection is crucial since the Outpainter creates a diversity of completions. Using classifier entropy yields completions consistent with inputs, while the trained discriminator provides more detail. Combined selection produces both consistent and detailed generations.}
	\label{fig:selection}
\end{figure}

\vspace{2mm}
\noindent {\bf Qualitative Results.} An autoregressive approach alone does not fully explain our success. As we examine in Figure \ref{fig:selection},
the variety of autoregressive completions means sample selection is critical.
While classifier entropy \cite{razavi2019generating} selects sensible completions, they tend to lack detailed texture (left).
In contrast, a discriminator selects completions with realistic textures, but they may not make sense with the entire scene (middle).
We find the selection methods are complimentary. Combined they select sensible completions with realistic detail (right).

\vspace{2mm}
\noindent {\bf Quantitative Results.} Table \ref{tab:ablations} confirms the qualitative results. 
On RealEstate10K, the baseline classifier and trained discriminator perform about as good or better than a single completion.
Again, combining the discriminator and classifier yields the best selections.
Combining is also helpful on Matterport.
However, its scanned environments tend to result in more homogeneous lighting, in comparison with the 
reflection effects of light in real images.
As a result, a single completion is typically sufficient to maximize autoregressive performance. 
Finally, in the table, we confirm outpainting in a VQ-VAE space is superior to using RGB.

\begin{table}[t!]
\caption{{\bf Synthesis Ablations}. Comparison of autoregressive model and selection criteria.
Our method beats all ablations on RealEstate10K. 
On Matterport, the same selection trends are true. 
However, Matterport's scanned environments exhibit homogeneous lighting, so a single completion is sufficient to maximize autoregressive performance.
}
\centering 
\resizebox{\ifdim\width>\linewidth
        \linewidth
      \else
        \width
      \fi}{!}
{
  \begin{tabular}{l c c c c c c}
    \toprule
    Method & \multicolumn{2}{c}{Matterport} & \multicolumn{2}{c}{RealEstate10K}\\

     & A/B vs. & FID $\downarrow$ & A/B vs. & FID $\downarrow$ \\
		& Ours $\uparrow$ &  & Ours $\uparrow$ & \\
\midrule
RGB Autoregressive & 41.3\% & 60.73 & 29.6\% & 31.90 \\ 
1 Completion & 52.3\% & 55.46 & 38.4\% & 28.04  \\
Classifier Selection & 47.7\% & 59.78 & 44.9\% & 28.71   \\
Discriminator Selection & 47.9\% & 56.49 & 47.7\%  & 26.30   \\
\midrule
Ours & - & 56.36 & - & 25.53    \\
\bottomrule
\end{tabular}
\vspace{-0.1in}
}
\label{tab:ablations}
\end{table}

\begin{figure}[t!]
	\centering
			{\includegraphics[width=\linewidth]{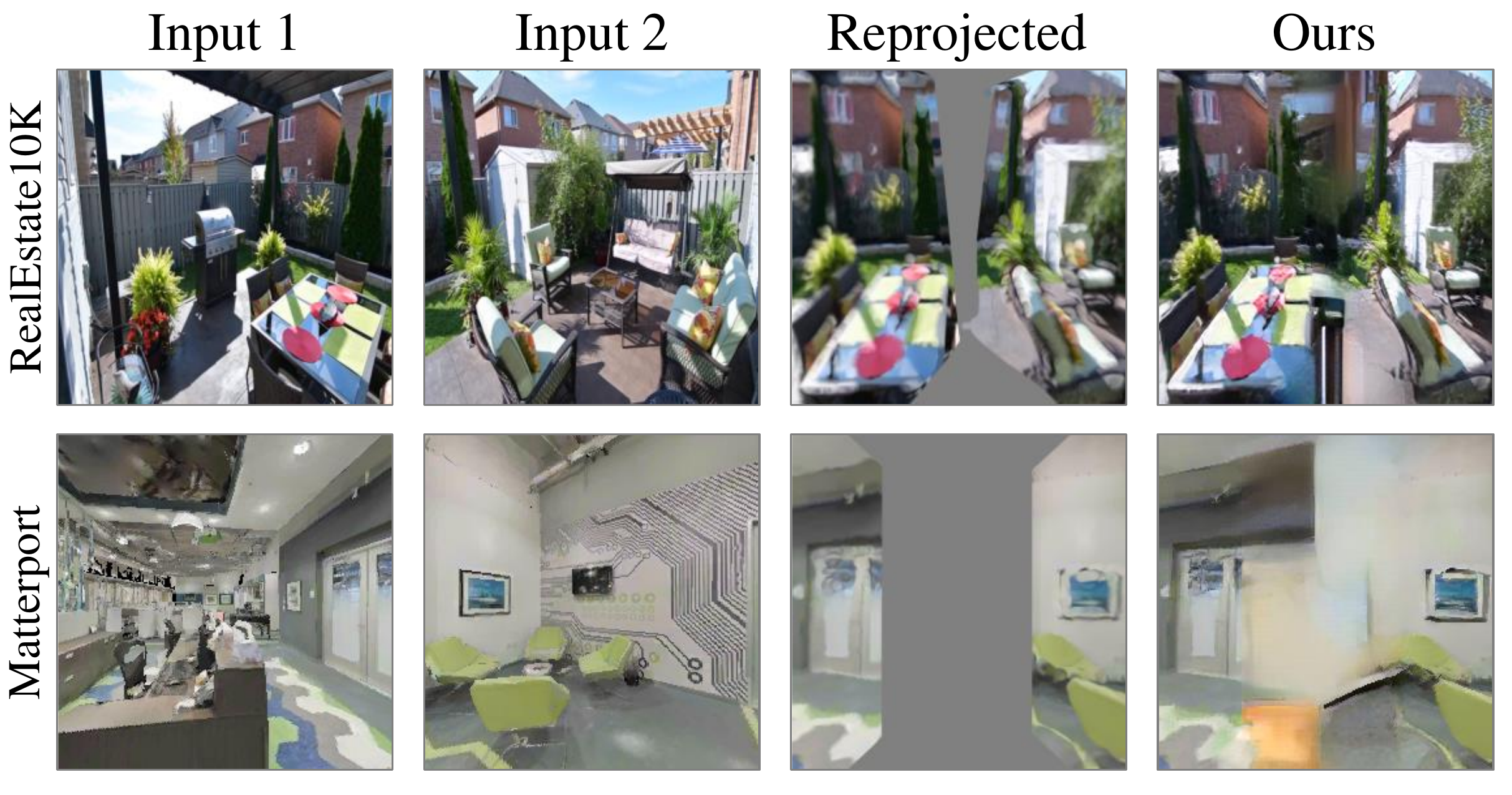}}
    \captionof{figure}{{\bf Two Input Synthesis.} The proposed method can readily generalize to two input images due to building on point clouds.}
	\label{fig:2input}
\end{figure}

\vspace{.2em}
\section{Discussion} \label{sec:discussion}

We see synthesizing a rich, full world from a single image as a steep new challenge.
Requiring only a single input opens up new experiences, but even with a single image, we see
3D awareness as important for good results and generality. For instance, our model's
3D awareness enables the application of our system to two views as shown in Figure~\ref{fig:2input} by ingesting
two point clouds.

\vspace{2mm}
\par \noindent {\bf Acknowledgments.} We thank Angel Chang, Richard Tucker, and Noah Snavely for allowing us to share frames from their datasets, and Olivia Wiles and Ajay Jain for easily extended code.
Thanks Shengyi Qian, Linyi Jin, Karan Desai and Nilesh Kulkarni for the helpful discussions.

\appendix
\newpage
\section{Appendix}

Video results available on the paper website give a thorough sense of model quality and consistency. 
As stated in the paper, the proposed method
tends to produce high-quality, consistent {\it scenes}. In contrast, 
baselines such as SynSin -- 6X are unable to create content, 
and ablations such as No 3D Accumulation are wildly inconsistent.
These results are best seen in video;
and are available at this URL:
\href{https://crockwell.github.io/pixelsynth/#video_results}{https://crockwell.github.io/pixelsynth/$\#$video$\_$results}

The pdf portion of the supplemental material shows:
detailed descriptions of model architectures (Section \ref{sec:architecture});
implementation details (Section \ref{sec:details});
details about the experimental setup (Section \ref{sec:setup});
additional results (Section \ref{sec:results});
and additional and A/B testing details
(Section \ref{sec:annotation}).

\subsection{Model Architecture}
\label{sec:architecture}

As stated in the paper, a forward pass of the model takes in a single image and produces a consistent image in a novel view $\pB$. This involves
multiple components: a depth module $D$ that maps images to depthmaps (producing point clouds); a projector $\pi$ that projects a point
cloud to a novel view; an outpainter $O$ that can outpaint missing pixels by autoregressive modeling on the latent space of a VQ-VAE;
and a refinement $R$ module that adds details and corrects mistakes on a full image.
We now provide more architectural details for each component; source code is available at \href{https://github.com/crockwell/pixelsynth}{https://github.com/crockwell/pixelsynth}.

\vspace{2mm}
\par \noindent {\bf Depth Module $D$:} The Depth Module takes in a $256 \times 256 \times 3$ image and predicts a depth for each pixel, yielding
a $256 \times 256 \times 1$ depthmap. For fair comparison, we follow the U-Net used in SynSin \cite{wiles2020synsin}, which consists of 8 encoding blocks that are mirrored by 8
decoding blocks.

\vspace{1mm}
\par \noindent {\it Encoder:} Each encoder block consists of a convolution (size: $4 \times 4$ / stride 2 / padding 1) followed by BatchNorm and leaky ReLU (negative slope of $0.2$). Each block halves
the width and height. The first convolution has 32 filters (mapping $3$ channels to $32$ channels); filter counts double at each block until reaching 256; they then remain constant.

\vspace{1mm}
\par \noindent {\it Decoder:} The decoder mirrors the encoder. Each block consists of a ReLU, $2\times$ bilinear upsampling, convolution ($3\times 3$ / stride 1 / padding 1), and BatchNorm (except
the last layer). Mirroring the encoder, filter counts remain the same ($256$) until the feature map has been upsampled to $\frac{1}{16}$th of the input size. Filter counts then start halving. The last block does not have BatchNorm
and has a final tanh at the top of the network.

\vspace{2mm}
\par \noindent {\bf Projector $\pi$:} The Projector takes a colored point cloud and pose $\pB$ and projects it as if seen at $\pB$. This produces a $256 \times 256 \times 3$ image along with an indication that 
of which pixels were projected to and which need to be outpainted.
We implement the projection with the point cloud rendering functions from Pytorch3D \cite{ravi2020accelerating}. Our design decisions follow SynSin \cite{wiles2020synsin} for fair comparison: We alpha-composite points in the z-buffer and
accumulate within a radius of 4 pixels.

We find that one change is important for autoregressive outpainting:
we do not consider reprojected pixels at the edge of the visible region. Pixels that fall just outside the projected
point cloud's silhouette can be non-zero: each rendered pixel is a function of the projected points within a radius, and points just outside the silhouette are
the result of interpolating some points and the missing regions. If we do not remove this, autoregressive outpainting begins with a border that is the mean color, which it tends
to continue. We prevent this by treating border pixels as background/to-be-outpainted.

\vspace{2mm}
\par \noindent {\bf Outpainter $O$:} The Outpainter takes as input the $256 \times 256$ reprojection from the Projector, possibly including large missing regions, and outpaints to a full image. This consists of performing autoregressive
outpainting on the latent space of an autoencoder. Crucially, the autoregressive model follows an image-specific order because each image and new pose yields different missing regions.
We describe the autoencoder, followed by the autoregressive model, then the autoregressive order. Our design decisions aim to make lightweight versions of VQVAE2 \cite{razavi2019generating} for the autoencoder and the PixelCNN++ \cite{Salimans2017PixeCNN}
used in Locally Masked Convolutions \cite{jain2020locally} for the autoregressive model.

\vspace{1mm}
\par \noindent {\it Autoencoder:} We follow a lightweight adaptation of VQVAE2 \cite{razavi2019generating} that maps a $256 \times 256 \times 3$ input to a $32 \times 32 \times 1$ quantized embedding space $Z$ and back. The encoder consists of first 3 convolution blocks followed by 2 ResNet blocks, then 2 convolution blocks and 2 ResNet blocks.
The encoder produces a $32 \times 32 \times 64$ continuous output; each pixel in the encoded continous space is quantized to an embedding $Z_{i,j,1} \in \mathbb{Z}^{512}_{1}$. 
The decoder first upsamples using a transpose convolution followed by a convolution. 
Then, it mirrors the encoder with 2 ResNet blocks followed by 2 transpose convolution blocks to produce a $256 \times 256 \times 3$ output.

\vspace{1mm}
\par \noindent {\it Autoregressive model:} The autoregressive model is a
lightweight PixelCNN++ \cite{Salimans2017PixeCNN} that produces, as output, a distribution
over the 512 possible quantized embedding values. Every convolution in the
network uses locally-masked convolutions \cite{jain2020locally} for custom
completion ordering. We follow the general design used by Locally Masked
Convolutions \cite{jain2020locally} on CIFAR-10 consisting of 30 Gated ResNet blocks with 160 filters. However,
we reduce its computational cost by using 12 Gated ResNet blocks with 80 filters, keeping everything
else constant. For more details, we refer the reader to \cite{jain2020locally}.

\vspace{1mm}
\par \noindent {\it Autoregressive ordering:} Autoregressive outpainting
follows an image-specific order. We must use this custom order because outpainting works best when
one predicts adjacent pixels in a sequence using as much known data as possible. In some scenarios
(e.g., extending a center crop), this can be achieved with a fixed order. However, in our case
the particular points that must be outpainted depend on the depthmap as seen in the first
image and the new pose from which it is projected.

Our order (Figure 4 of main paper) aims to go from closest out, following a spiral pattern. We achieve this by sorting the background/to-be-outpainted
pixels in ascending distance to the center of mass of the foreground/projected-to pixels. We start 
with the closest pixel and add the closest adjacent point not in the generation order. We repeat
this process until the entire image is ordering; ties due to pixels having equal distance are broken
using a spiral pattern outwards from the center of mass.

\vspace{2mm}
\par \noindent {\bf Refinement Module $R$:} The Refinement Module takes in a full $256 \times 256 \times 3$ outpainted image and produces the final, refined output of the same size $256 \times 256 \times 3$. This is
trained adversarially and so it consists of a generator and discriminator. 

\vspace{1mm}
\par \noindent {\it Generator:} The generator follows BigGAN \cite{brock2018large} and SynSin \cite{wiles2020synsin} and consists of 8 ResNet blocks capped with a tanh.
Following \cite{brock2018large}, there is an added downsampling block and with noise injection into BatchNorm throughout. Specifically, each ResNet block follows
the following structure, consisting of two paths which are added. The first path consists of: a linear layer that injects noise followed by BatchNorm; ReLU; convolution (size: $3 \times 3$ / stride 1 / padding 1);  
a linear layer to inject noise followed by BatchNorm; ReLU; and convolution (size $3 \times 3$ / stride 1 / padding 1). 
The second path consists as a convolution (size $1 \times 1$ / stride 1 / padding 0), which is added to the input.
Whenever a ResNet block downsamples, it uses average pooling to downsample the input during residual connection; whenever
it upsamples, it uses bilinear upsampling. 

\vspace{1mm}
\par \noindent {\it Discriminator:} The discriminator consists of 2 discriminator modules at different scales. Each discriminator contains 5 convolution blocks. Each block
contains a convolution (size $4 \times 4$ / stride 2), followed by a Leaky ReLU (negative slope 0.2). The middle three blocks additionally contain an instance normalization layer between the conv and leaky ReLU.

\subsection{Implementation Details}
\label{sec:details}

\par \noindent {\bf Outpainting Inference.} The Outpainter's autoregressive model produces its outputs by sampling. The forward pass produces a probability distribution over the vector embedded classes 
for every missing pixel in the $32 \times 32$ image. We find that best results are obtained by generating a set of full completions, followed by selection, and by adjusting
the sampling temperature used during inference to balance detail and error.

\vspace{1mm}
\par \noindent {\it Sample selection.} 
For each image, we generate 50 completions and select the best. We use a
combination of the discriminator loss from $R$ and a classifier entropy. The classifier
is trained on MIT Places 365 \cite{zhou2017places}. Selection uses the average
of ranks obtained by: (1) ranking in descending order of discriminator loss (since
higher loss tends to correspond to issues with details); (2) ranking in ascending order of
entropy (since sensible completions tend to be confident predictions of the classifier).

\vspace{1mm}
\par \noindent {\it Sampling Temperature.}
Sampling temperature is important for balancing diversity and error.
On Matterport, we use a sampling temperature of 0.5, which we find reduces strange completions but is still detailed.
On RealEstate10K, we also use 0.5 temperature for the 1-completion model. Again, we see this temperature best balances realism with detailed completions. On RealEstate10K's 50-completion model, we find we can increase sampling temperature to 0.7. While this means more completions are not sensible, we can select the best using an automated method. Therefore, the final outputs are more detailed and are still realistic.

\vspace{2mm}
\par \noindent {\bf Generating Multiple Viewing Directions.} To create scenes as approximated in Figure 1 and seen in the supplemental video, support views are synthesized in eight directions: up, left, down, right, up-left, up-right, down-left, and down-right.
These directions are selected to give a sense of all directions, and can be used to synthesize interior views without additional outpainting thereafter. 
When using multiple support views, we accumulate in a similar manner to the first support view: we
lift existing information to 3D, reproject into the new support view, and outpaint as needed. 
In other words, we do not outpaint the same region again.

\subsection{Experimental Setup}
\label{sec:setup}

As detailed in the paper, we use two datasets to evaluate. 
Matterport uses embodied agent navigation to select paired views, while RealEstate10K selects paired frames from real video clips.
We therefore use different processes to achieve a shared goal of selecting image pairs with large angle change.

\vspace{1mm}
\par \noindent {\it Matterport:} 
Matterport selection is straightforward because of embodiment. This consists of randomly drawing angle change in an embodied agent with a maximum of $120^\circ$ in each direction. We use a dataset of 3.6k pairs.

\vspace{1mm}
\par \noindent {\it RealEstate10K:} 
On RealEstate10K, it is harder to select large angle changes because we are instead selecting from pairs of images in real videos.
We select pairs of images such that the pairs have at least 20 degrees angle change.
In order to attain such pairs, we allow sampling from anywhere in video clips, which can be over 270 frames apart.
To minimize the number of view changes so extreme that input and target view are in different rooms, we limit translation at 1.0 meters, and angle at 60 degrees. 
We only consider pairs that meet this criteria, rather than resampling. Our filtered test set
chooses 3600 pairs from over 3.5 million possible pairs across over 2.4k video clips.
All selected evaluation pairs are cached for replicability.

\subsection{Additional Results}
\label{sec:results}

We report additional results, including video predictions, which provide the best simultaneous display of quality and consistency.

\vspace{2mm}
\par
\begin{figure*}[t!]
	\centering
			{\includegraphics[width=.9\textwidth]{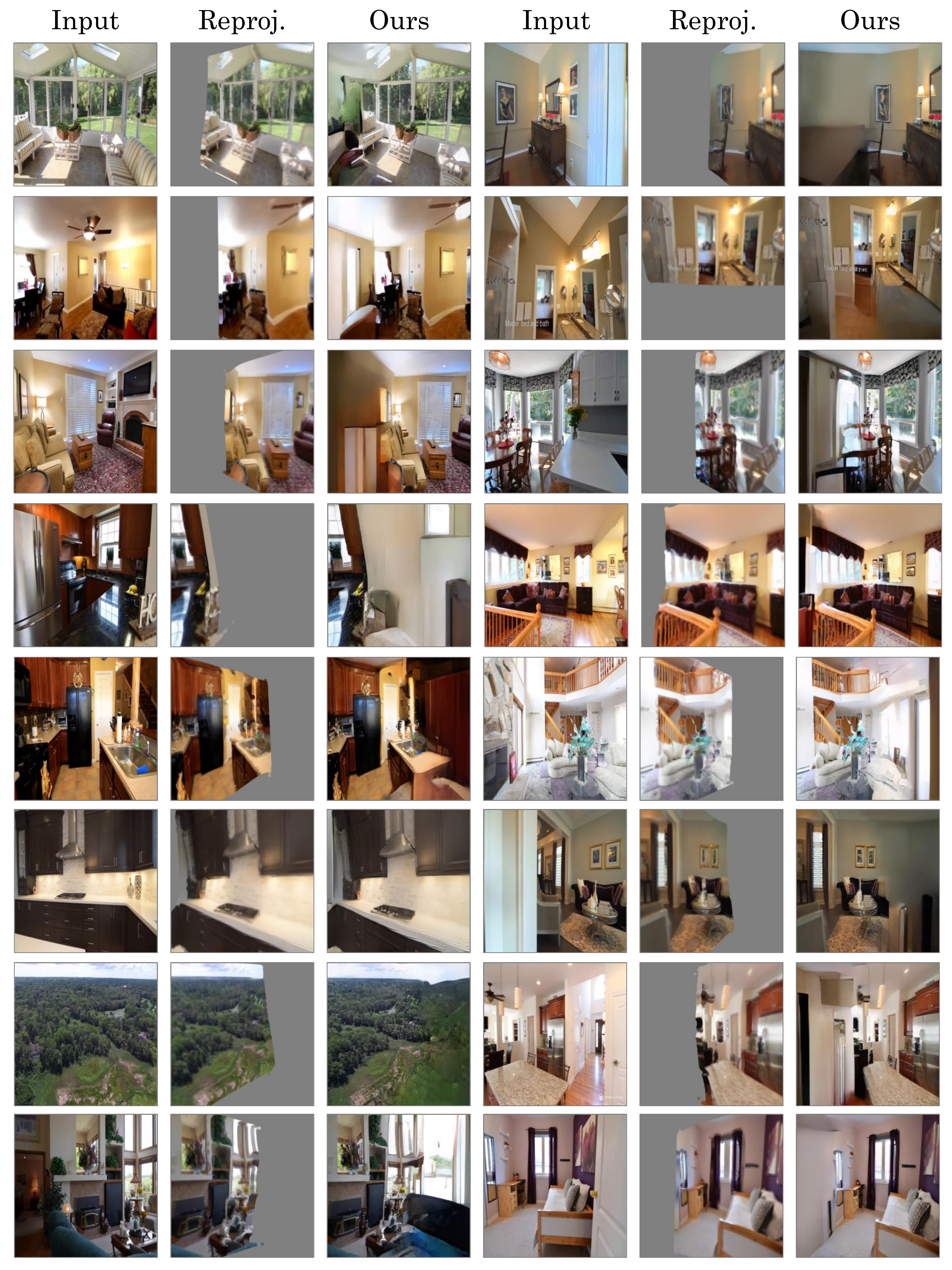}}
\captionof{figure}{{\bf Additional Results on RealEstate10K.}}
\label{fig:re}
\end{figure*}

\begin{figure*}[t!]
	\centering
			{\includegraphics[width=.9\textwidth]{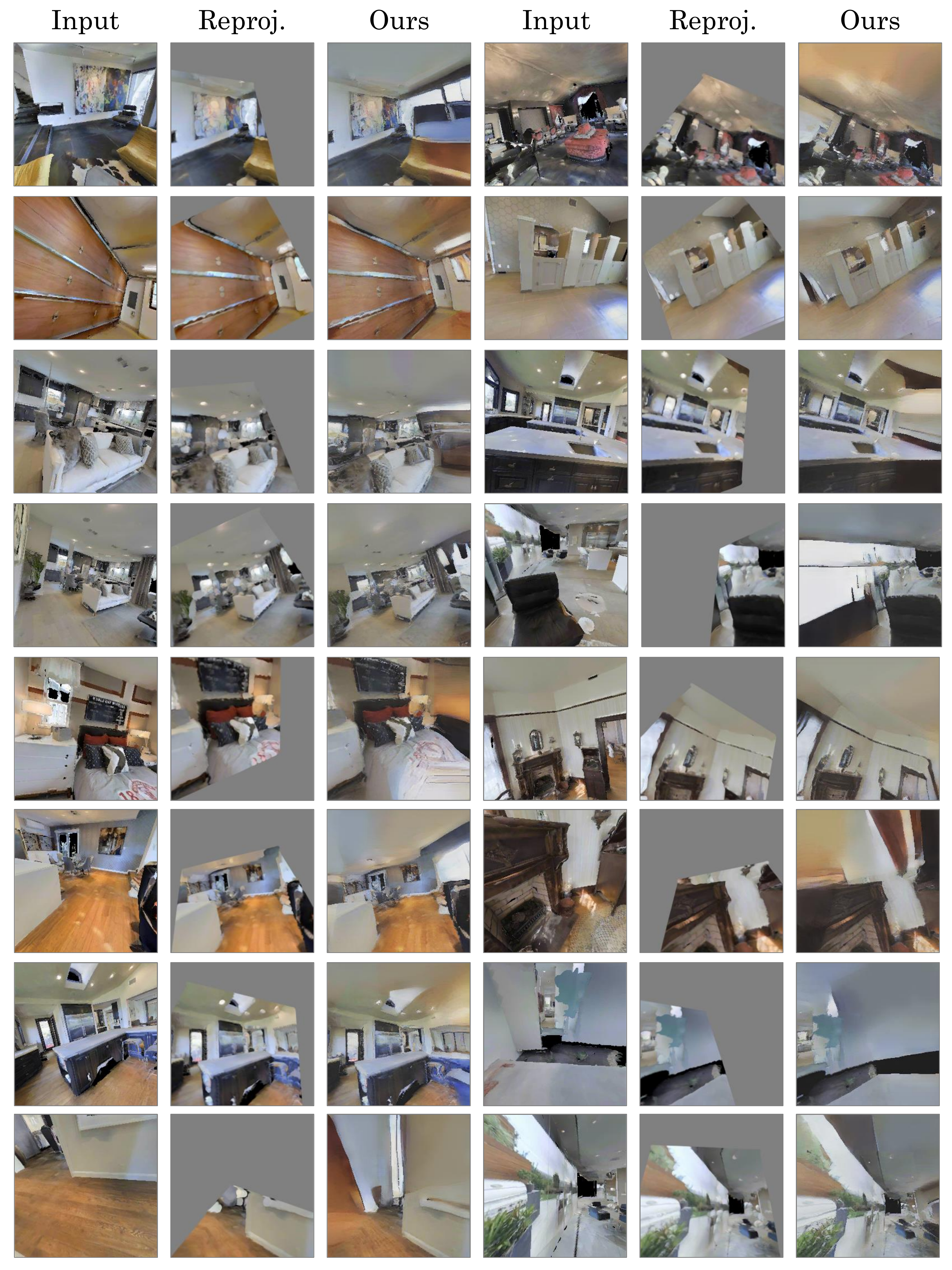}}
\captionof{figure}{{\bf Additional Results on Matterport.}}
\label{fig:mp3d}
\end{figure*}

 \noindent {\bf Additional Qualitative Results:} We first report additional qualitative results. Video results give a thorough sense of model quality and consistency, and are in the project webpage. As stated in the paper, the proposed method tends to produce high-quality, consistent scenes. In contrast, baselines such as SynSin - 6X are unable to create content, and ablations such as No 3D Accumulation are wildly inconsistent. Additional frame-level generated images are available in Figures \ref{fig:re} and \ref{fig:mp3d}.

\vspace{2mm}

\par \noindent {\bf Additional Quantitative Results:} We report more extensive results for generated image quality.
This is an expansion on Table 2 in the paper.
Although these automated metrics are poor measures for extrapolation, we present in more detail in Table \ref{tab:bad2} for completeness. 
Ground truth depth is available in Matterport, meaning visible and non-visible regions can be attained, similar to SynSin.
The same is not true of RealEstate10K, which contains real videos for which extensive ground truth labeling is prohibitive.

We reiterate the caution from the paper that PSNR has poor correlation with perceived quality when there are multiple possible completions
(for instance during outpainting): Appearance Flow is competitive with other methods on PSNR but loses to our proposed method $98\%$ of the time
in A/B testing.

\vspace{2mm}
\par \noindent {\bf Limitations:} Our primary limitation is outpainting both consistent and detailed content, especially on large view changes. 
There is a trade-off between the two, as a greater diversity of samples is required for detail, but can result in inaccurate content. 
While the approach of rejection sampling can improve outpainting errors, they remain a challenge, particularly on Matterport.
In fact, on Matterport we reduce sampling temperature to minimize inconsistent completions, which can result in less detail.
For instance, in Supp. Figure \ref{fig:mp3d} row 3 column 1, notice missing content tends to repeat visible content rather than ending visible objects and creating new ones.

\begin{table}[h!]
\caption{{\bf Full PSNR and Perc Sim:} Traditional metrics such as PSNR are poor measures for extrapolation tasks, but are reported for reference.}
\centering 
\resizebox{\ifdim\width>\linewidth
        \linewidth
      \else
        \width
      \fi}{!}
{
  \begin{tabular}{l c c c c c c}
    \toprule
    Method & \multicolumn{4}{c}{Matterport} & \multicolumn{2}{c}{RealEstate10K}\\
     & \multicolumn{3}{c}{PSNR $\uparrow$} & Perc Sim $\downarrow$ & PSNR $\uparrow$ & Perc Sim $\downarrow$ \\
 &  Both & InVis & Vis &  &  &  \\    
\midrule
Tatarchenko \textit{et al.} & 13.72 & 13.59 & 15.24 & 3.82 & 10.63 & 3.98 \\
Appearance Flow & 13.16 & 13.11 & 14.75 & 3.68 & 11.95 & 3.95 \\
Single-View MPI  & - & - & - & - & 12.73 & 3.45 \\
SynSin  & 15.05 & 14.35 & 17.86 & 3.13 & 13.92 & 2.77 \\
SynSin - Sequential & 14.31 & 13.36 & 17.65 & 3.14 & 13.30 & 2.78  \\
SynSin - 6X & 15.52 & 14.94 & 17.98 & 3.16 & 14.17 & 2.78 \\
SynSin - 6X, Sequential & 15.61 & 15.07 & 17.92 & 3.17 & 14.21 & 2.73 \\
\midrule
Ours & 14.60 & 13.58 & 18.08 & 3.17 & 13.10 & 2.88   \\
    \bottomrule
\end{tabular}
\vspace{-0.1in}
}
\label{tab:bad2}
\end{table}

\subsection{Additional and A/B Testing Details}
\label{sec:annotation}

A/B testing is the primary measure for success throughout experiments.
This is common in work on extrapolation, as automated metrics tend to struggle.
We detail our A/B testing framework below.

All A/B testing follows a standard A/B testing paradigm where human workers are shown images and are asked which is preferred given an input image. Workers were given instructions and example images and labels. They then had to pass a qualifier assessing whether they understood the task. Workers were also monitored by gold standard sentinel labels. All annotations were gathered using thehive.ai, a website similar to Amazon Mechanical Turk. Tasks are detailed below, and we share worker instructions.

\vspace{2mm}
\noindent {\bf Evaluating Quality via A/B:} For comparisons of quality, we use novel view synthesis. 
A/B testing presents workers with the input image reprojected into a new view, and asks them to select the final image that makes more sense given this image. 
The reason we use reprojections as worker input instead of input images is it makes the A/B testing much clearer for workers.
Using an input and rotation are very difficult to visualize, and thus difficult to compare across models.
We also do not compare to ground truth output, as final images can vary drastically from ground truth and still be highly reasonable.

\begin{figure*}[t!]
	\centering
			{\includegraphics[width=.95\textwidth, height=.95\textheight,keepaspectratio]{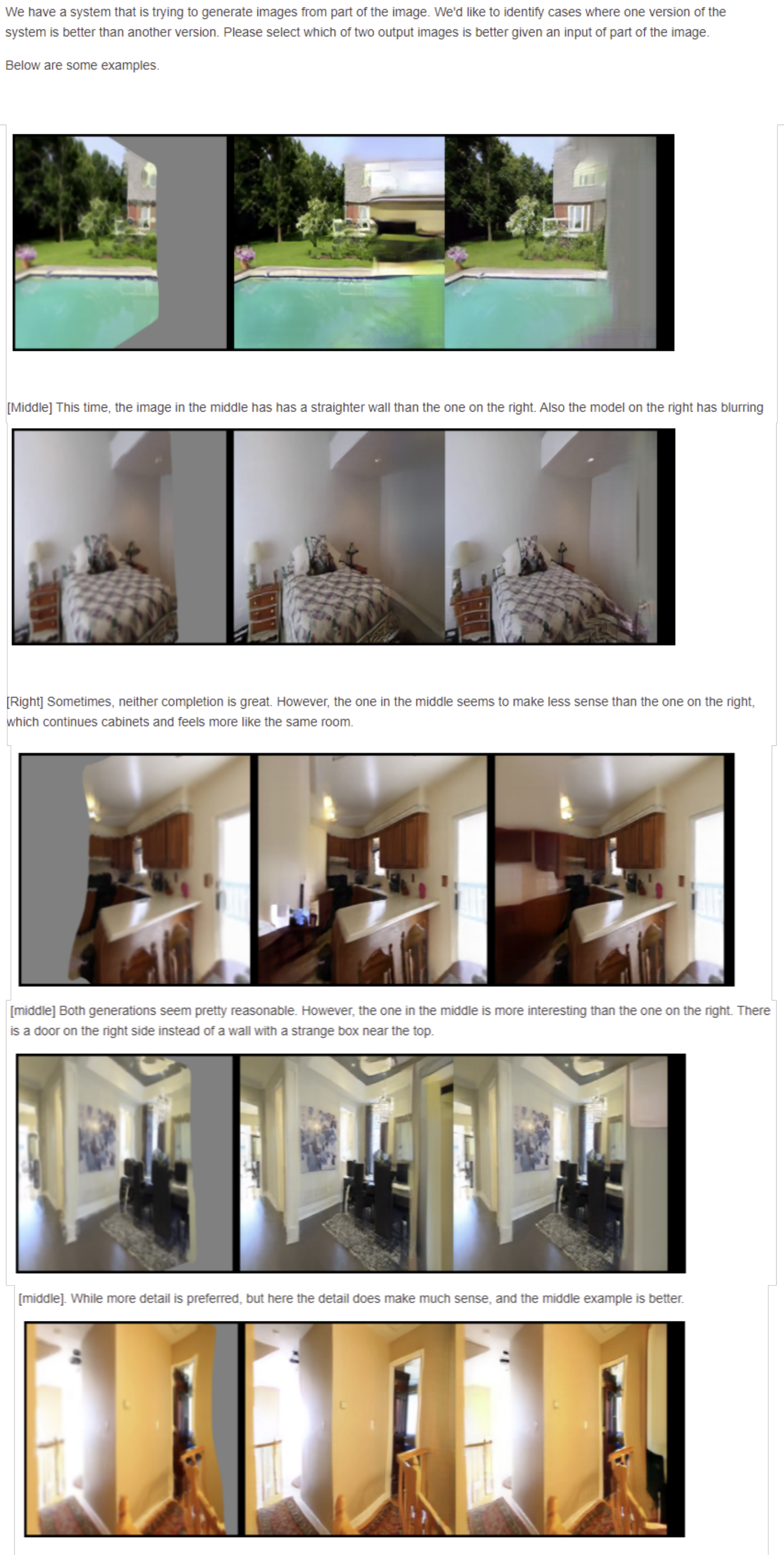}}
\captionof{figure}{{\bf Quality A/B Worker Instructions.}}
\label{fig:quality}
\end{figure*}

Instructions are shared in Figure \ref{fig:quality}. Ties are not allowed; final selection requires agreement of at least two workers.
Reprojections use the learned depth from our model, which is effective on large view changes and therefore tend to be accurate, compared to ground truth images.

\vspace{2mm}
\noindent {\bf Evaluating Consistency via A/B:} For consistency A/B testing, we use a similar novel view synthesis setup.
However, instead of predicting one image, the model predicts two images such that the second generated image is half the rotation and translation of the first.

\begin{figure*}[t!]
	\centering
			{\includegraphics[width=.95\textwidth, height=.95\textheight,keepaspectratio]{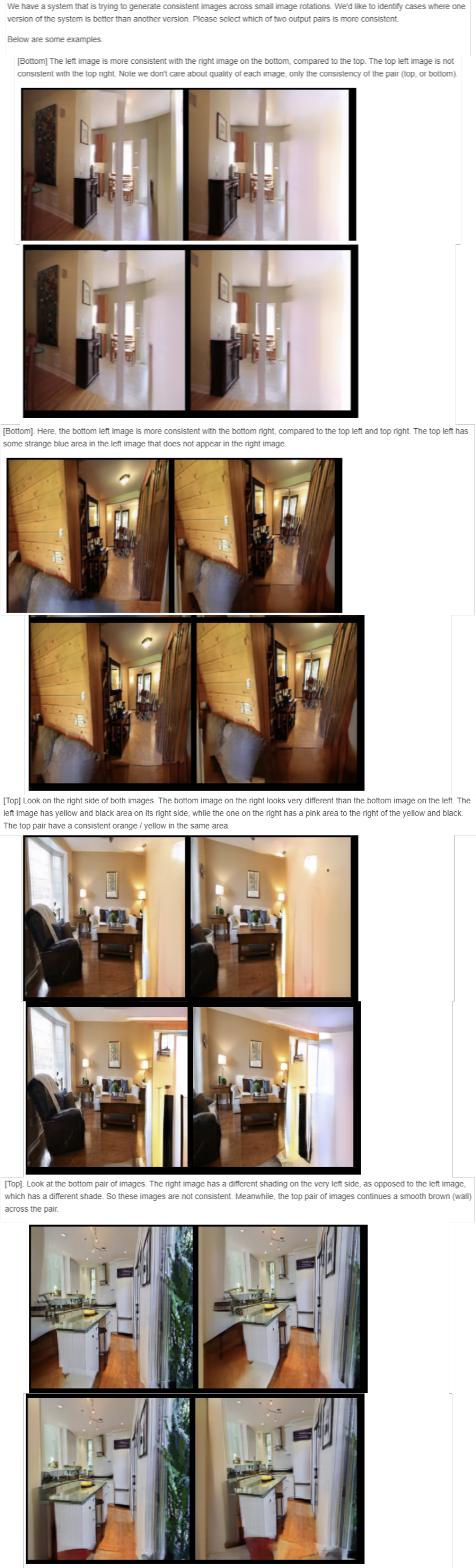}}
 \captionof{figure}{{\bf Consistency A/B Worker Instructions.} }
\label{fig:consistency3}
\end{figure*}

Workers are then asked to compare generated image pairs across methods.
We ask them to do so by showing them a pair of images generated by two competing methods, pairs being stacked vertically. 
We do not use the input or ground truth images as the goal is not to judge quality, but only to judge consistency.
Thus, even if one pair looks less realistic, it should be selected. Full instructions are displayed in Figure \ref{fig:consistency3}.

Consistency can be difficult for workers to judge as it requires attending to small regions of each pair of images that may be slightly different.
We therefore take several steps to maximize worker success.
We use fixed rotations of large size ($\sim$ $35^\circ$ horizontal, $\sim$ $17.5^\circ$ vertical) to ensure angle change is apparent.
We also constrain the rotation so that it must be in the horizontal and vertical directions, as additionally using roll rotations can make transformations confusing.
Movement is also limited to that related to embodied rotation, since movement opposing rotation can make consistency difficult to evaluate.
Finally, rotation is randomly selected for each image from one of eight possible directions seen in Figure 1: up, left, down, right, up-left, up-right, down-left, and down-right.
This allows us to explicitly specify rotation direction of each image pair to help workers attend to specific regions of image pairs.

\vspace{2mm}
\noindent {\bf Evaluating Consistency via Homography:}
We validate A/B consistency using PSNR and Perceptual Similarity via homography. We use the same setup as in A/B testing, but instead apply only pure rotations to images.
This enables homographies to warp across generated images. We do so in each pair both from intermediate to extreme images and from extreme to intermediate images. 
PSNR is then calculated on overlapping regions, while Perc Sim is calculated on warped images with non-overlapping regions masked.
Scores are averaged across both directions in each pair.

\clearpage


\end{document}